\documentclass{article}

\usepackage[preprint]{neurips_2022}

\usepackage{setspace}
\usepackage[utf8]{inputenc} 
\usepackage[T1]{fontenc}    
\usepackage[pagebackref=true,breaklinks=true,colorlinks,bookmarks=false]{hyperref}       
\usepackage{url}            
\usepackage{booktabs}       
\usepackage{amsfonts}       
\usepackage{nicefrac}       
\usepackage{microtype}      
\usepackage{xcolor}         

\usepackage{graphicx}

\usepackage{tikz}
\usepackage{comment}
\usepackage{amsmath,amssymb} 
\usepackage{color}

\usepackage[accsupp]{axessibility}  

\usepackage{multirow}
\usepackage{adjustbox}
\usepackage{subcaption}
\usepackage[labelfont=bf]{caption}
\usepackage{overpic}
\usepackage{bm}
\usepackage{colortbl}
\usepackage{diagbox}
\usepackage{mathtools}
\definecolor{maroon}{cmyk}{0,0.87,0.68,0.32}

\usepackage{multirow}
\usepackage{array}
\newcommand{\PreserveBackslash}[1]{\let\temp=\\#1\let\\=\temp}
\newcolumntype{C}[1]{>{\PreserveBackslash\centering}p{#1}}
\usepackage{tikz}

\usepackage{pifont}
\usepackage{subfloat} 
\usepackage{bbding}
\usepackage{overpic}

\title{Multi-Camera Calibration Free BEV Representation for 3D Object Detection}

\author{%
  Hongxiang Jiang$^{1,2}$\thanks{This work is done when Hongxiang Jiang is an intern at Horizon Robotics.}
  \quad 
  Wenming Meng$^2$ 
  \quad 
  Hongmei Zhu$^2$ 
  \quad 
  Qian Zhang$^2$ 
  \quad 
  Jihao Yin$^1$\thanks{Jihao Yin (jihaoyin@buaa.edu.cn) is the corresponding author with the School of Aerospace, Beihang University, Beijing, China.}
  \quad 
  \vspace{.5em} 
  \\
  $^1$Beihang University 
  \quad 
  $^2$Horizon Robotics
  \vspace{.5em} 
}
\begin{document}

\maketitle

\begin{abstract}
In advanced paradigms of autonomous driving, learning Bird's Eye View (BEV) representation from surrounding views is crucial for multi-task framework. However, existing methods based on depth estimation or camera-driven attention are not stable to obtain transformation under noisy camera parameters, mainly with two challenges, accurate depth prediction and calibration. In this work, we present a completely Multi-Camera \textbf{C}alibration \textbf{F}ree \textbf{T}ransformer (CFT) for robust BEV representation, which focuses on exploring implicit mapping, not relied on camera intrinsics and extrinsics. To guide better feature learning from image views to BEV, CFT mines potential 3D information in BEV via our designed position-aware enhancement (PA). Instead of camera-driven point-wise or global transformation, for interaction within more effective region and lower computation cost, we propose a view-aware attention which also reduces redundant computation and promotes converge. CFT achieves $49.7\%$ NDS on the nuScenes detection task leaderboard, which is the first work removing camera parameters, comparable to other geometry-guided methods. Without temporal input and other modal information, CFT achieves second highest performance with a smaller image input ($1600 \times 640$). Thanks to view-attention variant, CFT reduces memory and transformer FLOPs for vanilla attention by about $12\%$ and $60\%$, respectively, with improved NDS by $1.0\%$. Moreover, its natural robustness to noisy camera parameters makes CFT more competitive.
\end{abstract}

\section{Introduction}
\begin{figure}[htb]
	\centering  
	\subfloat[Explicit Camera-Driven]{
	    \label{compare(a)}
		\includegraphics[width=0.33\linewidth]{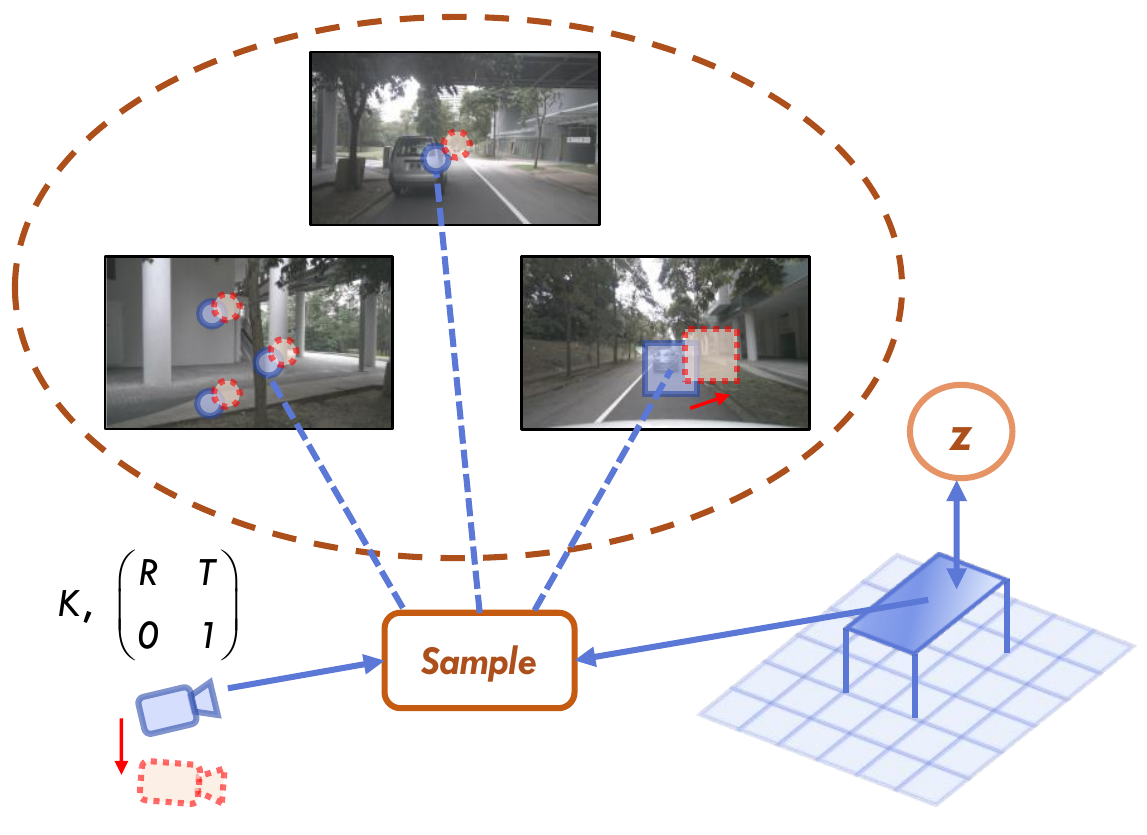}}
	\subfloat[Implicit Camera-Driven]{
	    \label{compare(b)}
		\includegraphics[width=0.33\linewidth]{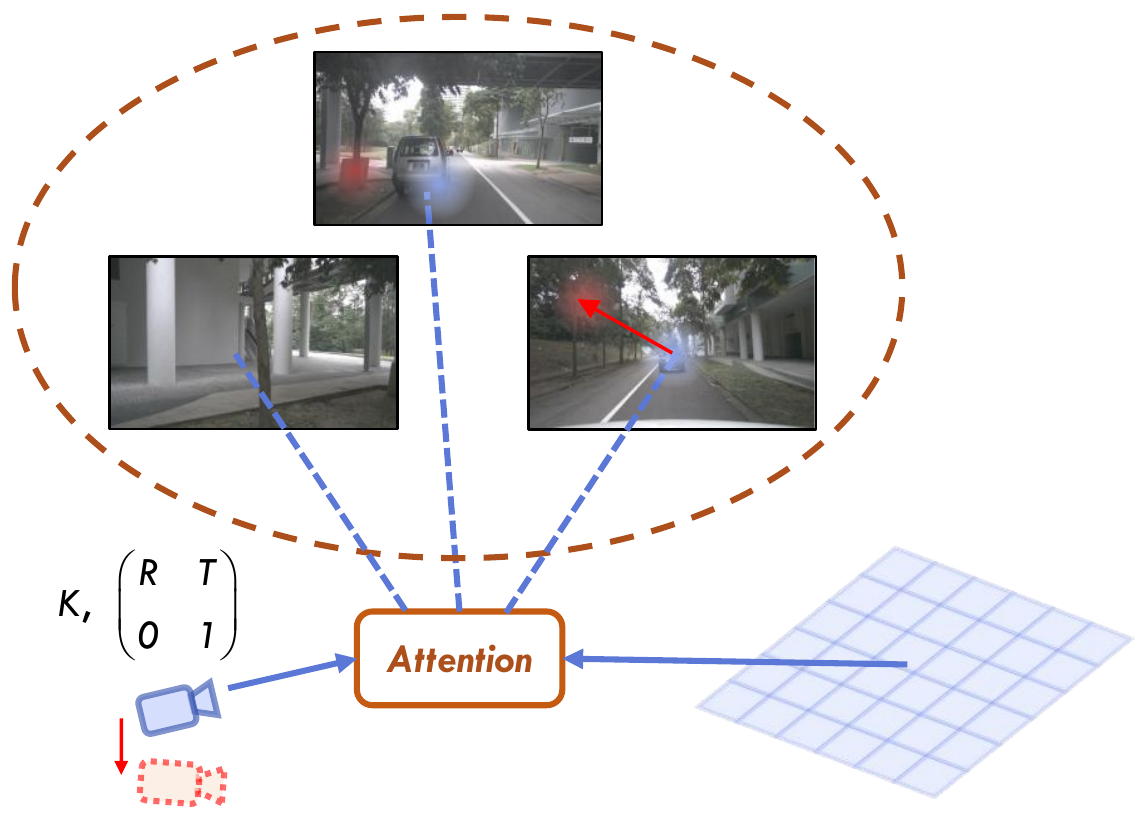}}
	\subfloat[Calibration Free]{
		\label{compare(c)}
		\includegraphics[width=0.30\linewidth]{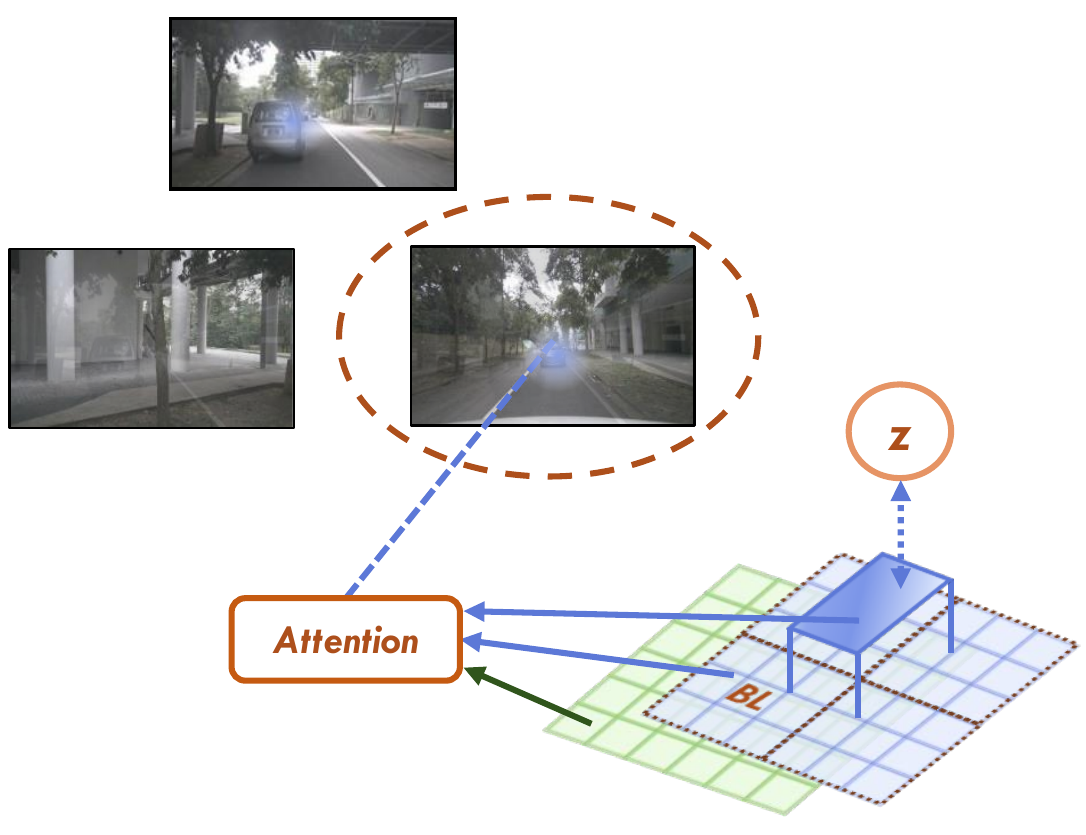}}
	\caption{\textbf{Illustration of our calibration free BEV representation.} (a) and (b) show explicit or implicit camera-driven attention interaction, $z$ in (a) is the true height coordinate corresponding to the predicted value of each BEV grid. Bias in camera parameter affects the learning of correct transformations. (c) is the proposed CFT, where the green and blue grids represent BEV content embedding and positional embedding, respectively, divided into 2$\times$2 windows for view-level attention. The additional embedding of $z$ could be obtained in a supervised or unsupervised manner, then restructured together with other embedding for attention.}
\end{figure}

3D object detection from multi-camera 2D images is a critical perception technique for autonomous driving systems with compared to expensive LiDAR-based \cite{chen2017multi,lang2019pointpillars,bewley2020range} or multi-modal approaches \cite{Wang_2021_CVPR,vora2020pointpainting,yoo20203d,yin2021multimodal,bai2022transfusion,nabati2021centerfusion,ijcai2022p116}. Recent approaches emphasize transforming 2D image features to sparse instance-level \cite{chen2022polar, shi2022srcn3d, Wang2021DETR3D3O} or dense Bird's Eye View (BEV) representation \cite{Huang2021BEVDetHM, Li2022BEVFormerLB,liu2022petrv2}, characterizing the 3D structure of the surrounding environment. Although some depth-based detectors \cite{Huang2021BEVDetHM,jiang2022polarformer,li2022bevdepth,liu2022petrv2,zhang2022beverse} incorporate depth estimation to introduce such 3D information, the extra depth supervision is acquired for preciser detection. Therefore, other paradigms \cite{Li2022BEVFormerLB,Wang2021DETR3D3O} directly learn the transformation based on the attention mechanism \cite{vaswani2017attention}. They explicitly predict the 3D coordinates of queries. With these coordinates, the camera intrinsic and extrinsic parameters, local image features could be sampled to perform cross-attention, as shown in Fig. \ref{compare(a)}. However, the obtained representation is not stable due to the inaccurate prediction, calibration error or dynamic variety of the camera parameters in real-world scenes. Inference speed also suffers from point
-wise coordinate projection and sampling. 

To learn representation more stably, researchers \cite{Liu2022PETRPE, Zhou_2022_CVPR} indirectly discover the relationship of different views by implicitly generating camera-driven 3D positional embedding. This bypasses the position estimation, but attention computation is still sensitive to camera parameters, which is shown in Fig. \ref{compare(b)}. In fact, the existing analysis\cite{Li2022BEVFormerLB,liu2022petrv2,philion2020lift} also shows that the noisy extrinsics have a great impact on the 
results. Besides, global attention between each position of the BEV grids and that of the image pixels causes high computational cost. Consequently, considering the above problems with explicit or implicit camera-driven attention, removing geometry prior for robust BEV representation and reaching a good balance between performance, speed and computational cost is of great importance.

In this work, we propose a multi-camera \textbf{C}alibration \textbf{F}ree \textbf{T}ransformer (CFT), learning BEV representation for 3D object detection without any geometric guidance. Fig. \ref{compare(c)} briefly illustrates our CFT, which helps the camera parameters-free translation between image views and BEV via implicitly enhancing 3D feature in the BEV embedding. Specifically, CFT proposes a content-based position-aware enhancement (PA). Thanks to PA that captures location and content separately, even implicit unsupervised learning can reliably extract 3D positional information, similar to the results of adding explicit supervision in PA. The original 2D coordinates encoding of BEV could contain richer 3D cues, gaining sufficient interaction with image pixels. This not only stably learns the features of the objects for correct perception, but also enables faster inference.

Committed to save computational cost of our proposed CFT model, we further present a lightweight view-aware attention (VA) instead of geometry-guided sparse sampling, simultaneously applied to the intra- and inter-view transformation. In addition to effectively reducing FLOPs and memory, view-attention ignores redundant computations, which promotes faster learning of attention. Compared with global attention, our built different schemes of view selection on BEV all achieve better results .

Overall, our contributions could be summarized as follows:
\begin{itemize}
\item We propose CFT to learn robust BEV representation for multi-camera 3D object detection without camera parameters. It is the first work that achieves comparable or even superior performance than geometry-dependent methods in terms of accuracy and speed ($45.5\%$ NDS, $3.6$ FPS vs $44.8\%$ NDS, $2.7$ FPS).
\item We design PA in obtaining richer 3D information of BEV, which could fully mine height features from content embedding, promoting the establishment of relationships between image views and BEV.
\item VA is presented for reducing redundant computation. In comparison with global transformation, the FLOPs of the attention and the memory is reduced by about $60\%\sim78\%$ and $12\%\sim17\%$ in VA. Additionally, it promotes better interaction within the effective region.
\item Experiments on nuScenes detection task evaluate that our proposed CFT achieves $49.7\%$ NDS (Oct. 5th, 2022), competitive compared to the state-of-the-art methods without temporal or other modal input (50.4\% NDS) on the leaderboard. In particular, CFT has no degradation compared to other models for noisy extrinsics.
\end{itemize}
\section{Related work}

Multi-camera 3D object detection predicts the 3D bounding boxes of the objects of interest from the input surrounding views. Motivated by typical works in 2D detection \cite{carion2020end,zhu2020deformable,sun2021sparse}, researchers combine 3D prior and propose different 3D object detection frameworks to directly achieve sparse object-level features extraction \cite{Wang2021DETR3D3O,chen2022polar,shi2022srcn3d,chen2022graph}. In recent new paradigms of autonomous driving, the BEV space attracts much attention because of its advantages in perception\cite{reading2021categorical,yang2021projecting,zou2022hft,chen2022persformer,philion2020lift}, prediction \cite{Akan2022ECCV,hu2021fiery}, multi-task learning \cite{xie2022m,zhang2022beverse,lu2022learning,can2021structured} and downstream planning \cite{philion2020lift}, etc. Thus, some advanced methods perform 3D object detection with obtained representation on the BEV space. BEVDet and CenterPoint \cite{yin2021center} perform keypoint-based detection \cite{zhou2019objects} by locating the object center point on the map view. PETR and BEVFormer utilize the end-to-end 3D detection head in DETR \cite{carion2020end}, treating all BEV grids as dense queries. For detecting occluded and dynamic objects better, multi-frame temporal information is also introduced into the origin detection architecture \cite{huang2022bevdet4d,liu2022petrv2,qin2022uniformer}.

\begin{table*}[htb]
    \caption{\textbf{Comparison of related approaches.} All of them are related to multi-camera 3D object detection or BEV representation learning. $"\checkmark"$ in camera-driven means that depth-based methods use camera parameters, while not further divided into explicit or implicit.
}
	\centering
	\setlength{\tabcolsep}{1.3mm}{
		\linespread{1.1}\selectfont
		\begin{tabular}{ccccc}  \toprule
			Method      &Depth? &Camera-Driven?  &BEV Representation? &Transformation Type?\\  \hline
            BEVDet      &\checkmark             &\checkmark                   &\checkmark            &Point-wise\\
            BEVDepth      &\checkmark             &\checkmark                   &\checkmark            &Point-wise\\
			DETR3D       &\ding{55}              &Explicit                  &\ding{55}            &Point-wise\\
			PolarDETR    &\ding{55}             &Explicit                  &\ding{55}              &Point-wise\\
			BEVFormer    &\ding{55}             &Explicit                  &\checkmark    &Point-wise\\
            GKT          &\ding{55}                    &Explicit                  &\checkmark    &Point-wise\\
			PETR         &\ding{55}              &Implicit
            &\checkmark    &Global\\
			CVT          &\ding{55}             &Implicit                  &\checkmark    &Global\\
			\hline
			Ours         &\ding{55}                   &\ding{55}        &\checkmark    &View\\
			\hline
	\end{tabular}}
    \label{related models}
\end{table*}

To learn BEV representation from surrounding views, depth-based methods \cite{philion2020lift,rukhovich2022imvoxelnet,li2022bevdepth, Huang2021BEVDetHM,wang2022mv,reading2021categorical} infer depth in image views and project them to the BEV plane with the extrinsics and intrinsics, where Unsupervised depth estimation remains challenges. Although BEVdepth \cite{li2022bevdepth} improves feature extracting and downstream tasks performance compared with other paradigms, additional supervision is a key issue, relatively more difficult to obtain. In contrast, inspired by the transformer framework for 2D object detection \cite{zhu2020deformable,carion2020end}, some works such as BEVformer \cite{Li2022BEVFormerLB, jiang2022polarformer,peng2022bevsegformer,chen2022efficient} emphasizes directly learning the transformation relationship between image view and BEV based on the attention mechanism. Specifically, they explicitly preset height or supervise the predicted height, then perform point-wise attention on BEV by sampling local image features according to the extrinsics and intrinsics. End-to-end representation learning and less computation are their advantages, but the reliance on 3D priors affects the robustness of the models, such as height errors or biases in camera parameters. Besides, coordinate transformation and sampling cause less efficient processing. Without the projection operation, PETR \cite{Liu2022PETRPE} and CVT \cite{Zhou_2022_CVPR} implicitly encodes the 3D information into image or BEV feature. But precise camera parameters are still necessary, which are indirectly used to guide the attention. Also, introducing global transformation increases the computational cost and optimization difficulty. Unlike previous methods, our proposed CFT does not estimate depth and completely removes camera parameters. Inspired by advanced vision transformers \cite{meng2021-CondDETR,liu2022dabdetr,li2022dn}, CFT decouples the positional and content embedding in the position-aware enhancement, and further mines richer 3D information, thereby effectively learning stable BEV representations. Instead of point-wise attention with camera guidance or redundant global attention, a view-attention is presented  to reduce the computational and accelerates the establishment of transformation relations. 

For the various related approaches mentioned above, we summarize the main differences between some representative ones and our proposed CFT in the Tab. \ref{related models}.
\section{Method}
\label{method}
\begin{figure*}[htb]
\centering
\includegraphics[width=\textwidth]{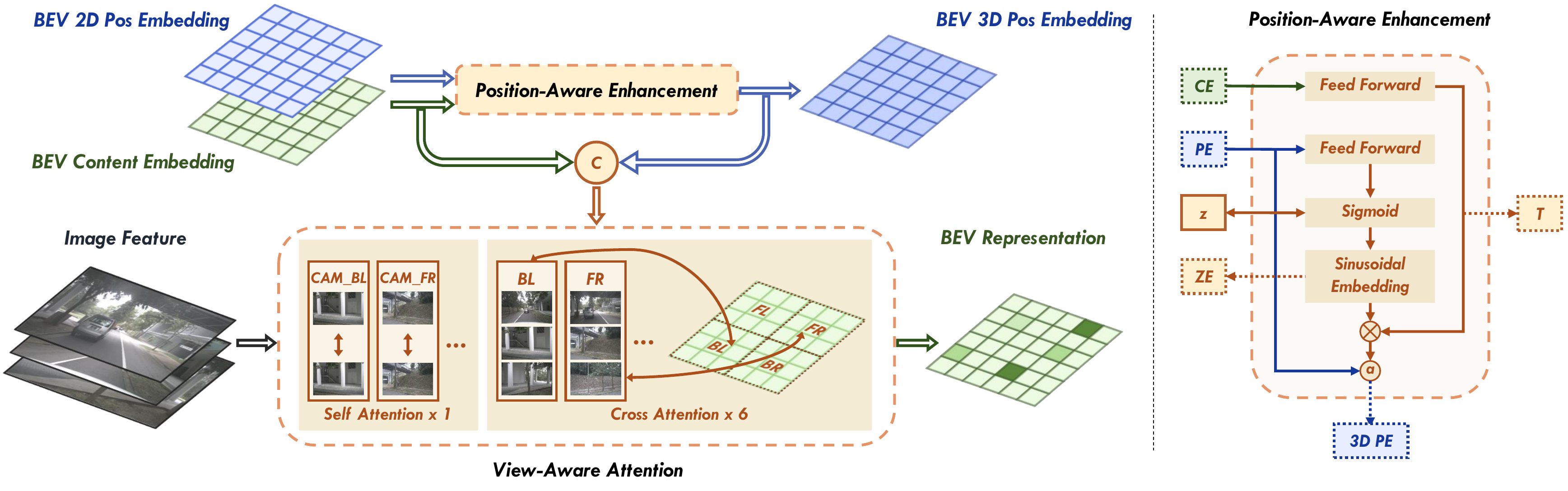}
\caption{\textbf{Our CFT for BEV representation learning.} PA introduces separate content and 2D positional embedding for BEV feature, and the height embedding, ZE, is extracted from them, which is illustrated in the right part of the figure. With the encoding reorganization, ($\otimes$, @, and $c$ denotes dot, add, concatenation, respectively), the enhanced BEV feature is input to VA to interact with image features. Both self-attention and cross-attention are modulated in VA based on the view prior. Each view image in self-attention is computed only with itself, where $CAM\_BL$, $CAM\_FR$, etc. represent different surrounding views. In cross-attention, the BEV plane is divided to windows in various scheme, and $BL$, $FR$, etc. correspond to different parts. The image view groups are designed for each part to achieve more efficient calculation and accelerated convergence. After stacking multiple attention blocks, CFT completes the transformation of image views feature to BEV representation.
}
\label{figure2_CFT}
\end{figure*}

\subsection{Overall Architecture}
In this work, we focus on attention-based 3D object detection without introducing temporal information and propose a entire architecture which thoroughly eliminates the influence of the camera parameters. We first begin with the multi-view images input, $I \in \mathbb{R}^{N_v \times H \times W \times 3}$, where $N_v$ represents the number of views. Each view image is fed into a backbone network and neck layers for extracting multi-scale feature maps. In particular, we adopt the lowest scale feature map $F_s \in \mathbb{R}^{N_v \times H_s \times W_s \times C_s}$ for subsequent processing, i.e. only with the single-scale. Then $F_s$ interacts with the BEV embedding for obtaining  the BEV representation in our CFT, as shown in Fig. \ref{figure2_CFT}. CFT consists of two main parts: position-aware enhancement (PA) and view-aware attention (VA), which are utilized to capture BEV 3D positional information and perform lightweight attention, respectively. Finally, the prediction of the 3D bounding boxes is completed on the BEV space with the help of the designed detection head.

\subsection{Position-Aware Enhancement}
To capture 3D positional information in BEV without camera parameters, we start with common BEV 2D coordinates embedding $Q_p \in {\mathbb{R}^{H_b\times W_b\times{C_p}}}$ and further update it, which assigns a unique feature encoding to each BEV grid $(h, w)$. `ally, we first roughly infer the reference height $z_{ref}$ for $(h, w)$ on the basis of $Q_p$, and perform sinusoidal positional encoding \cite{vaswani2017attention} as follows:

\begin{equation}
\label{eq:1}
\begin{aligned}
    Q_{ref} = Sinusoidal(z_{ref}),\
    z_{ref} = Norm(Sigmoid(FFN(Q_p))),
\end{aligned}
\end{equation}
where $Norm$ scales the output of the activation layer to a preset height range, $Q_{ref}$ means the reference height encoding. To adjust the height adaptively for different objects, we further introduce the BEV content embedding $Q_c \in {\mathbb{R}^{H_b\times W_b \times{C_s}}}$. Then the matrix $M$ is learned based on $Q_c$ for refining the object height, and obtaining the updated results. It can be described as

\begin{equation}
\label{eq:2}
\begin{aligned}
    Q_{ep} = add(M \cdot Q_{ref}, Q_p),\ M = FFN(Q_c).
\end{aligned}
\end{equation}

\begin{figure}[htb]
	\centering  
	\subfloat[Implicit]{
	    \label{tsne(a)}
		\includegraphics[width=0.23\linewidth]{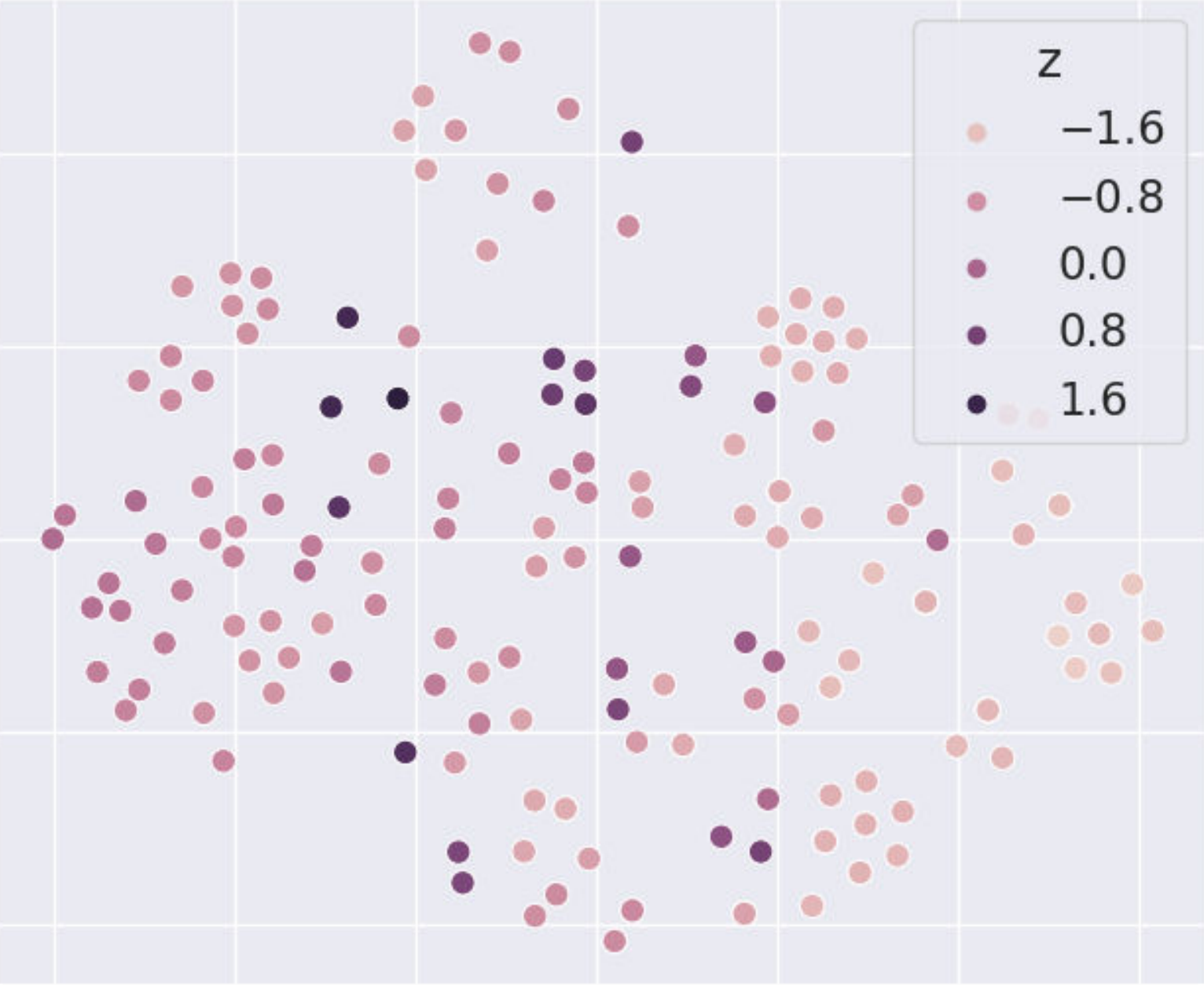}}
	\subfloat[Explicit]{
	    \label{tsne(b)}
		\includegraphics[width=0.23\linewidth]{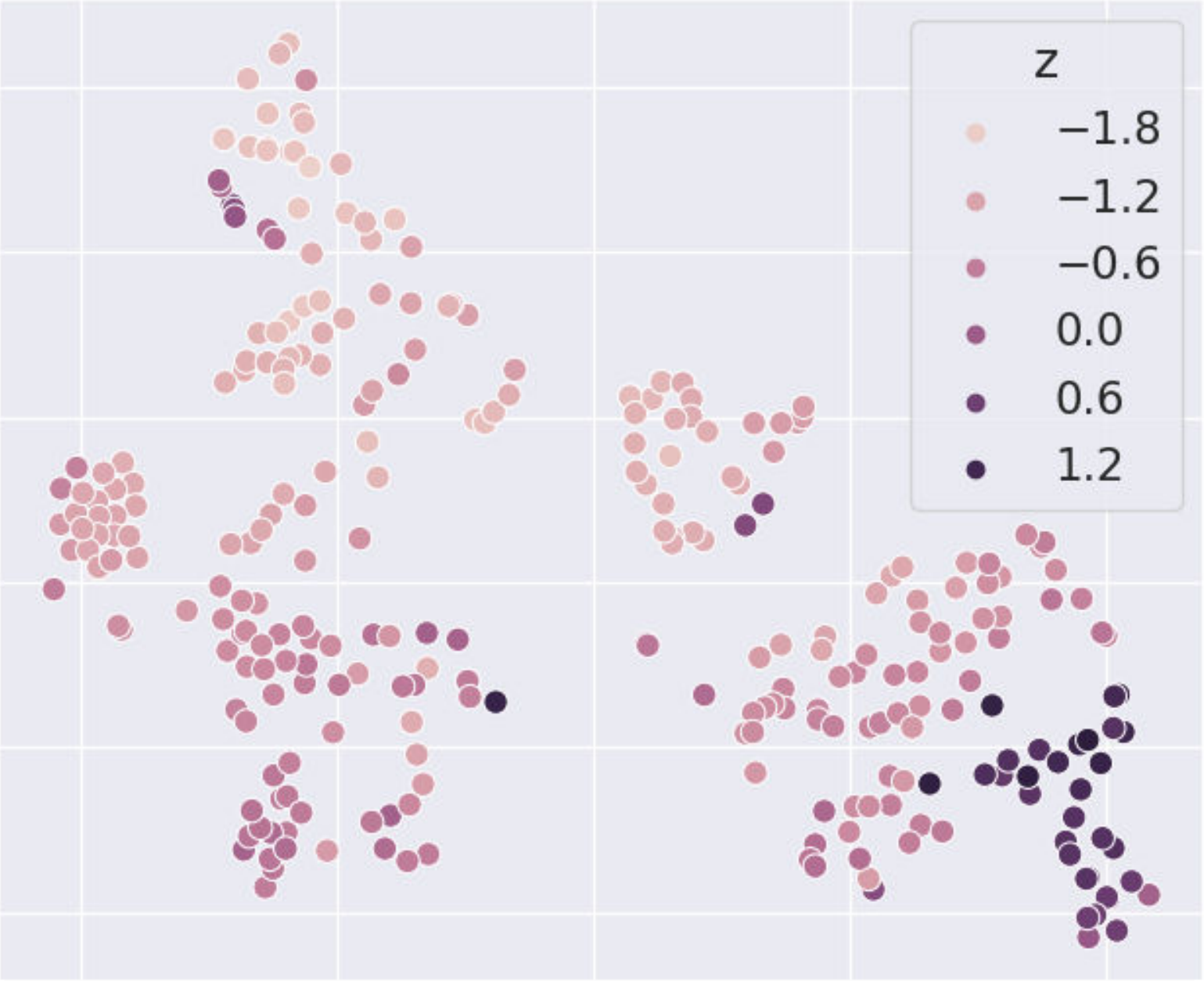}}
	\subfloat[Enhanced Implicit]{
		\label{tsne(c)}
		\includegraphics[width=0.465\linewidth]{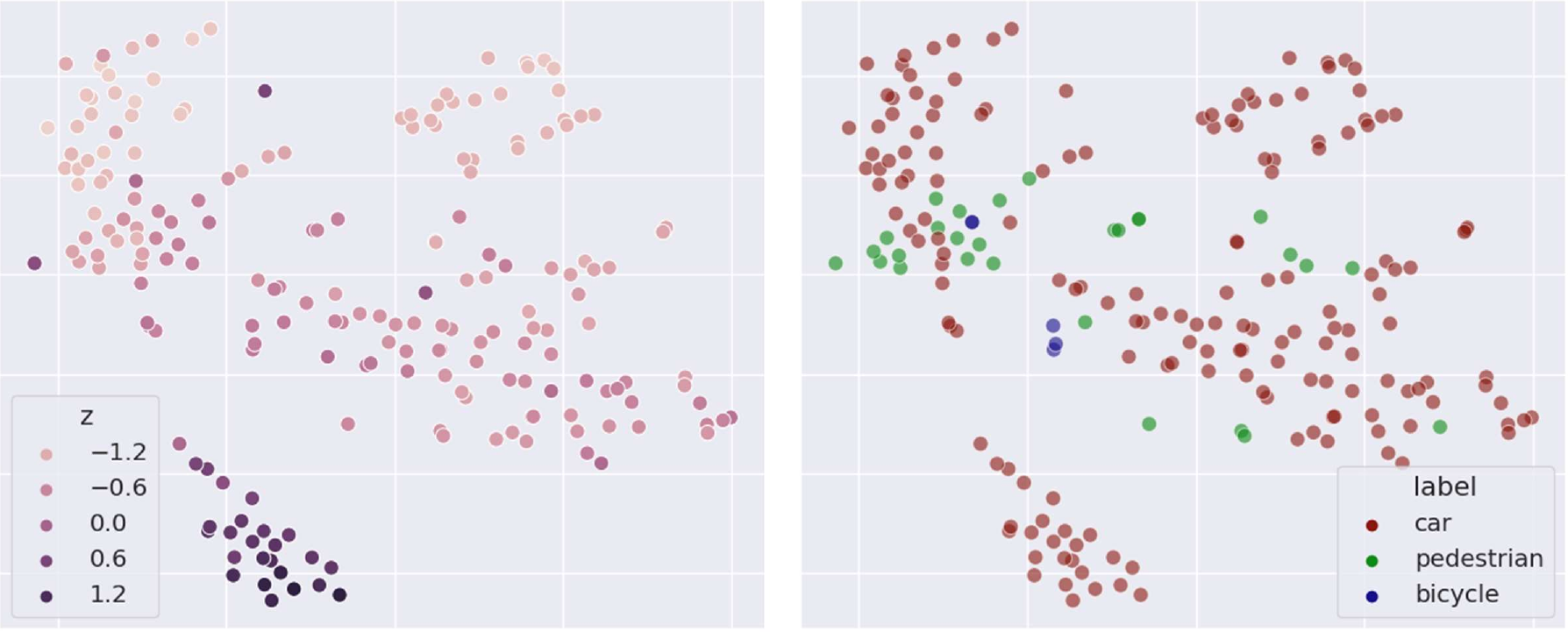}}
	\caption{\textbf{Relationship between z, labels and learned position embedding of objects.} The visualization is implemented by t-sne, a common feature dimension reduction method. Scatters in the figure represent the feature of the BEV grids where the true objects are located. In the left of Fig. \ref{tsne(c)}, different colors denote the classes of objects corresponding to these features. In other figures, the shades of color correspond to their height variations. Fig. \ref{tsne(b)} and \ref{tsne(c)} show more separability in height relative to \ref{tsne(a)}, with a more continuous change in color. In particular, labels are marked in Fig. \ref{tsne(c)} as well, further demonstrating that rather than the category information, the height information are extracted from the content for refinement. In other words, on the same category, we can still learn corresponding embedding according to different heights.}
	\label{tsne}
\end{figure}

This operation could extract position-related information in the content, and obtain $Q_{ep} \in {\mathbb{R}^{H_b\times W_b\times{C_p}}}$, which enhances the original insufficient 2D coordinates embedding. The whole step is shown in Fig. \ref{figure2_CFT}. Considering that vanilla attention mixes the content and position, which is not beneficial for our added $Q_{ep}$ to correctly learn position information, we restructure the two parts both in images and BEV:

\begin{equation}
\label{eq:3}
\begin{aligned}
    P_{image} &= cat(F_s, P_x, P_y, P_v), \\
    P_{bev} &= cat(Q_c, Q_{ep}),
\end{aligned}
\end{equation}

where $cat$ means the channel concatenation. The view encoding $P_v$ indirectly takes into account the relative relationship of the cameras. $P_{image}$ and $P_{bev}$ are used to generate $Q$, $K$, $V$ for attention. The above design clarifies the concept of content and position and ensures that position information is included in the refinement of the height. Even without introducing camera parameters, the sufficient interaction between $Q_{ep}$ and $P_x,P_y,P_z$ could guide the implicit learning of the coordinates mapping. In fact, the reference height $z_{ref}$ in PA comes directly from $Q_p$, tending to learn the overall distribution of the data, so adaptive refinement helps to more accurately represent the true object height. Additionally, though PA adds supervision for learning $z_{ref}$ with $L1$ loss by default, after removing the above supervision, the refined height embedding still exhibits the ability on height representation. 

Fig. \ref{tsne} demonstrates the feature visualization via t-sne \cite{vandermaaten08a}, which compares the relationship between the refined height encoding and height ground truth under different designs. In details, Fig. \ref{tsne(a)} represents only using 2D positional information and vanilla attention to learn the transformation, Fig. \ref{tsne(b)} and \ref{tsne(c)} respectively indicate whether there is supervision for predicting $z_{ref}$ in our PA. Regardless of the supervision, our PA helps the enhanced positional embedding of BEVs more correlated with height, whose height feature exhibits better separability compared to the simple design. Therefore, thanks to PA, camera extrinsics and intrinsics could be completely removed, still ensuring that 3D information is effectively learned.

\subsection{View-Aware Attention}
In standard global attention, $P_{image}$ and $P_{bev}$ will be flattened for calculation directly, excluding the channel dimension. Different from this computationally  redundant method with high cost, our proposed view-level attention variant only considers valuable interactions, shown in Fig. \ref{figure2_CFT}. We first reshape the image encoding into $P_{image}^{'} \in {\mathbb{R}^{N_v\times H_sW_s\times{C}}}$ to modulate self-attention, where $C = C_s + C_p$. Therefore, for each image, self-attention only involve $H_sW_s$ pixels in the same view. Then, we divide $N_v$ windows along the spatial dimension on BEV encoding $P_{bev}$. In fact, there are multiple division methods, bringing different benefits in terms of computation reduction and performance improvement. We choose the strategy with the best balance, which will be further analyzed theoretically and experimentally in Sec. \ref{Ablation studies}. 

Taking a rectangular division as an example, our method reshape $P_{bev}$ into $P_{bev}^{'} \in {\mathbb{R}^{N_v\times \frac{H_b}{N_H}\cdot\frac{W_b}{N_W}\times{C}}}$, where $N_v={N_H}{N_W},N_H=2,N_W=2$. It covers four BEV windows, $FL$, $FR$, $BL$, and $BR$ in Fig. \ref{figure2_CFT}. To preserve effective information, instead of performing cross-attention calculation on $P_{image}^{'}$ and $P_{bev}^{'}$, we continue to divide the view groups $G \in {\mathbb{R}^{N_v\times N_gH_sW_s\times{C}}}$ from $P_{image}^{'}$. Points from different windows on the BEV feature will interact with different view groups. The specific division is as follows:
\begin{equation}
\label{eq:4}
\begin{aligned}
    G_{FL}&=cat(P_{CAM\_FL}^{'},P_{CAM\_F}^{'},P_{CAM\_BL}^{'}),\ 
    G_{FR}=cat(P_{CAM\_F}^{'},P_{CAM\_FR}^{'},P_{CAM\_BR}^{'}),\\
    G_{BL}&=cat(P_{CAM\_FL}^{'},P_{CAM\_BL}^{'},P_{CAM\_B}^{'}),\ 
    G_{BR}=cat(P_{CAM\_FR}^{'},P_{CAM\_B}^{'},P_{CAM\_BR}^{'}),
\end{aligned}
\end{equation}
where different P' correspond to six views front-left, front, front-right, back-left, back, back-right respectively. Accordingly, for the grids of $BL$ window in $P_{bev}^{'}$, we choose view group $G_{BL}$ to contribute keys, where the content part contributes values for cross-attention, other windows perform similar calculations. After multiple layers of cross-attention, we obtain the final BEV representation $F_B \in {\mathbb{R}^{H_b \times W_b \times{C_s}}}$.

Our attention variant introduces view priors without relying on the extrinsics and intrinsics. In addition to reducing memory cost, it also removes redundant information via roughly considering the camera range, which is beneficial to learn BEV feature faster and better. VA is inherently robust to camera parameters and leads to more efficient inference than geometry-guided attention.

\subsection{Detection Head}
We upsampled the BEV expression $F_B$ by a factor of $4$ to obtain a large resolution BEV feature map $F_B^{'} \in {\mathbb{R}^{4H_b \times 4W_b \times{C_s}}}$, which is beneficial for small object detection. For 3D object detection task on $F_B^{'}$, we adopt the single-stage CenterPoint \cite{yin2021center} as detection head to predict the center heatmap with focal loss and regression map with $L1$ loss. Additionally, we make several adjustments to further simplify the training pipeline. We remove the structure of separate heads for different classes and regression terms, and set all Gaussian kernels to a fixed size when generating the ground truth. During inference, we utilize scale-nms \cite{Huang2021BEVDetHM} for decoding and post-processing of bounding boxes. 

\newcommand{\lack}[1]{\textcolor{cyan}{#1}}
\newcommand{\bug}[1]{\textcolor{brown}{#1}}
\newcommand{\es}[1]{\textcolor{magenta}{#1}}

\section{Experiment}

\subsection{Datasets and evaluation metrics}

We conduct all the experiments on the nuScenes \cite{nuscenes2019}, a public large-scale autonomous driving dataset. The nuScenes dataset collects 1000 driving scenes with 20 second length and manually divides the training set, validation set and test set. Our results are all from the evaluation of the \textit{val} set or the \textit{test} set. We mainly use 6-views images with size $1600 \times 900$ and annotations of 3D bounding boxes provided in 40k keyframes. For the detection task, these annotations come from 10 categories with their respective detection ranges.

We evaluate our approaches according to the official evaluation metrics provided by nuScenes, including mean Average Precision (mAP), a set of true positive metrics that measure translation (ATE), scale (ASE), orientation (AOE), velocity (AVE) and attribute (AAE) errors. Finally, the comprehensive score, nuScenes detection score (NDS), is derived from the weighted sum of the above metrics:

\begin{equation}
\label{eq:6}
\begin{aligned}
    NDS = \frac{1}{10}\left[5mAP + \sum_{mTP} max(1 - mTP, 0)\right].
\end{aligned}
\end{equation}

\subsection{Implementation details}
We adopt ResNet-101 \cite{he2016deep} backbone, pretrained by FCOS3D \cite{wang2021fcos3d}, and FPN \cite{lin2017feature} neck to extract image feature for our CFT by default, where image size is $1600 \times 640$. For the outputs of ResNet, we choose the feature map of size 1/64 and dimension $C_s=256$. We set BEV content and positional embedding as $64 \times 64 \times 256$, corresponding to the perception ranges of [-51.2m, 51.2m] for the X and Y axis. In VA module, they are divided into 4 windows with the size of $32 \times 32$, and fed into 1 layer of self-attention and 6 layers of cross-attention. In the default detection head, we set the heatmap of each object to $9 \times 9$ Gaussian kernel to generate ground truth. During bounding boxes decoding, scale-nms is consistent with BEVDet.

Except the model for official testing is trained on 8 Telsa V100, others are run on 8 RTX3090 GPUs, with 24 epochs and a total batch size of 8. We choose the AdamW \cite{loshchilov2018decoupled} optimizer with a step learning rate policy, which drops the learning rate at 20 and 23 epoch by a factor of 0.1. Learning rate is set to $2 \times 10^{-4}$.

\begin{table*}[htb]
    \caption{\textbf{Comparison with state-of-the-art methods on nuScenes \textit{val} set}. BEVDepth-S is BEVDepth with $\#$ is trained with CBGS \cite{zhu2019class}. $\dagger$ means that the multi-scale feature maps are utilized. For the fair comparison of the inference speed, FPS is evaluated under two different devices, 1 RTX3090 and 1 Telsa V100, which are abbreviated as G and T to ensure the alignment of different methods.
}
	\centering
	\setlength{\tabcolsep}{0.8mm}{
		\linespread{1.1}\selectfont
		\begin{tabular}{c|c|c|ccccccc}  \toprule
			Method      &Size
			&NDS$\uparrow$  &mAP$\uparrow$   &mATE$\downarrow$   
			&mASE$\downarrow$   &mAOE$\downarrow$   &mAVE$\downarrow$   &mAAE$\downarrow$ &FPS$\uparrow$\\ 
			\hline
			DETR3D$\dagger$  &1600$\times$900   &0.425 &0.346 &0.773 &\textbf{0.268} &0.383 &0.842 &0.216 &2.0 (G)\\
			BEVFormer$\dagger$  &1600$\times$900   &0.448 &0.375 &0.725 &0.272 &0.391 &0.802 &0.216 &2.7 (G)\\
			BEVDepth-S  &1408$\times$512   &0.408 &\textbf{0.376} &0.659 &0.267  &0.543 &1.059 &0.335 &5.2 (G)\\
			CFT-BEV3D            &1600$\times$900   &\underline{0.445} &0.335 &0.716	&0.277 &\textbf{0.373} &\textbf{0.671} &0.187 &\textbf{3.6} (G)\\
			CFT-BEV3D            &1600$\times$640   &0.444 &0.334 &\textbf{0.715} &0.278 &0.374 &\textbf{0.689} &\textbf{0.177} &\textbf{5.0} (G)\\
			CFT-BEV3D            &1408$\times$512   &0.415 &0.307 &0.732 &0.276 &0.476 &0.713 &0.186 &\textbf{6.0} (G)\\
		    \hline
			DETR3D$\#\dagger$   &1600$\times$900   &0.434 &0.349 &0.716 &0.268 &0.379 &0.842 &0.200 &\ding{55}\\
			PETR$\#$           &1600$\times$900   &0.442 &\textbf{0.370} &0.711 &\textbf{0.267} &0.412 &0.834 &\textbf{0.190} &1.7 (T)\\
			CFT-BEV3D$\#$           &1600$\times$900   &\textbf{0.455} &0.343 &\textbf{0.651} &0.274 &\textbf{0.338} &\textbf{0.716} &\textbf{0.184} &\textbf{2.6} (T)\\
			\hline
	\end{tabular}}
    \label{val}
\end{table*}

\begin{table*}[htb]
    \caption{\textbf{Comparison with state-of-the-art methods on nuScenes \textit{test} set}. The underline indicates that our CFT-BEV3D achieves the second best performance (until 05 October, 2022).
}
	\centering
	\setlength{\tabcolsep}{1.3mm}{
		\linespread{1.1}\selectfont
		\begin{tabular}{c|c|c|cccccc}  \toprule
			Method      &Multi-Scale 
			&NDS$\uparrow$  &mAP$\uparrow$   &mATE$\downarrow$   
			&mASE$\downarrow$   &mAOE$\downarrow$   &mAVE$\downarrow$   &mAAE$\downarrow$ \\ 
			\hline
			DETR3D      &\checkmark  &0.479 &0.412 &0.641 &0.255 &0.394 &0.845 &0.133\\
			BEVFormer   &\checkmark  &0.495 &0.435 &0.589 &0.254 &0.402 &0.842 &0.131\\
			PETR        &            &0.504 &\textbf{0.441} &0.593 &0.249 &0.383 &\textbf{0.808} &0.132\\
		    BEVDet      &            &0.488 &0.424 &0.524 &\textbf{0.242} &\textbf{0.373} &0.950 &0.148\\
			CFT-BEV3D   &            &\underline{0.497} &0.416 &\textbf{0.518} &0.250 &0.390 &0.829 &\textbf{0.124}\\
			\hline
	\end{tabular}}
    \label{test}
\end{table*}

\subsection{Main Results}
\label{Main Results}

\subsubsection{nuScenes \textit{val} set}
We compare the proposed method with four state-of-the-art models, DETR3D, BEVFormer, BEVDepth and PETR, as shown in Tab. \ref{val}, where CFT-BEV3D represents our CFT model for the 3D object detection task. For a fair comparison, we input image views with different sizes to align other methods. Under the same conditions, CFT-BEV3D outperforms DETR3D and PETR by $2.1\%$ (or $2.0\%$ without CBGS) and $1.3\%$ on NDS, respectively. At a smaller size of $1600 \times 640$, CFT-BEV3D still exceeds DETR3D $1.9\%$. In particular, even if we do not use the multi-scale feature, we can still obtain a competitive NDS score ($-0.3\%$ vs BEVFormer) compared to the multi-scale methods. Compared to the advanced method with additional deep supervision, BEVDepth, our method is still $0.7\%$ NDS higher at the same resolution $1408 \times 512$. Notably, even though BEVdepth surpasses most methods with larger resolution on mAP and mATE due to its accurate depth estimation, other metrics is relative low, so inferior to our CFT-BEV3D. In terms of inference speed, CFT-BEV3D exhibits higher FPS across different input scales. Furthermore, unlike other methods whose performance degrades under noisy extrinsics, CFT-BEV3D is stable, which could be far superior to other methods. Noise analysis of camera parameters will be more discussed in Sec. \ref{Noisy Extrinsics Analysis}.

\subsubsection{nuScenes \textit{test} set}
Tab. \ref{test} shows a comparison of methods without temporal input and other modal information on the nuScenes detection task leaderboard. All the models are trained with VoVNetV2 \cite{lee2020centermask} backbone, pretrained by DD3D \cite{park2021pseudo}. The difference is that our CFT-BEV3D exploits a smaller ($1600 \times 640$) resolution input and obtains the comparable NDS score, which is the second best and $0.7\%$ lower than PETR. 

\begin{table}[htb]
    \caption{\textbf{Different embedding design}. "Implicit" means a naive design with 2D learned positional embedding for image views and BEV, and the embedding is mixed for vanilla global attention. "Explicit" and "Enhanced Implicit" means whether if there is explicit supervision of reference height in our PA.
}
	\centering
	\setlength{\tabcolsep}{1.3mm}{
		\linespread{1.1}\selectfont
		\begin{tabular}{c|c|cccc}  \toprule
			Method
			&NDS$\uparrow$  &mAP$\uparrow$   &mATE$\downarrow$   
			&mASE$\downarrow$   &mAOE$\downarrow$\\ 
			\hline
			Implicit &0.424	&0.320 &0.709 &0.278 &0.419\\
			Explicit   &\textbf{0.444}	&0.334 &0.715 &\textbf{0.278} &\textbf{0.374}\\
			Enhanced Implicit &0.440	&\textbf{0.340} &\textbf{0.700} &0.281 &0.409\\
			\hline
	\end{tabular}}
    \label{embedding}
\end{table}

Overall, all the experiments illustrate that our CFT has a good performance on 3D object detection, comparable to some state-of-the-art methods. It maintains robustness to camera parameter with different noise levels, and achieves a balance of speed and performance.

\subsection{Ablation studies}
\label{Ablation studies}
In this section, we follow the configuration without CBGS training and on the nuscene validation set. To emphasize the importance of obtaining a model robust to camera parameters, we additionally perform noisy extrinsics analysis, mainly involving PETR and BEVFormer that outperform our CFT-BEV3D in Sec. \ref{Main Results}.

\subsubsection{Embedding Designs}
When constructing embedding based on our designed PA to enhance position information, we conduct experiments with three strategies for comparison: implicit, explicit and enhanced implicit, mentioned in Sec.\ref{method}. The detailed explanations and results are shown in Tab. \ref{embedding}.

Compared with the learned embedding, our "Explicit" PA improves $2.0\%$ on NDS, in which the improvement of mAOE and mAP is obvious by $4.5\%$ and $1.4\%$. It shows that the additional 3D enhanced information helps each BEV grid better correspond to the true position, which is beneficial to capture the correct image features. They are presented in the map view, thereby improving the prediction of orientation and category. Specially, "Enhanced Implicit" PA is comparable to "Explicit" on NDS, only the mAOE is slightly worse. 

The above experiments demonstrate that thanks to the design of PA, 3D information can be properly learned and enhanced to guide better feature transformation from image views to BEV.

\subsubsection{BEV Window Division}

\begin{figure}[htb]
	\centering  
	\subfloat[Rec 2 $\times$ 2 \hspace{-1.2cm}]{
	    \label{window(a)}
		\includegraphics[width=0.288\linewidth ]{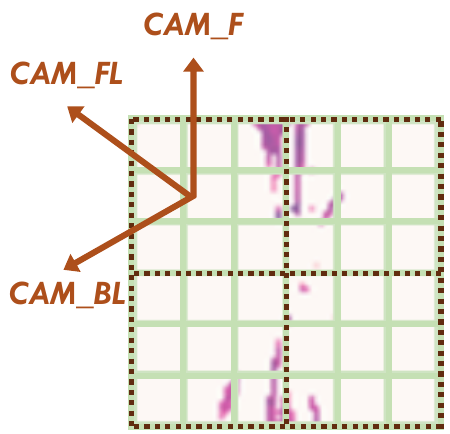} \hspace{-0.17cm}}
	\subfloat[Rec 2 $\times $3 \hspace{-0.05cm}]{
	    \label{window(b)}
		\includegraphics[width=0.21\linewidth]{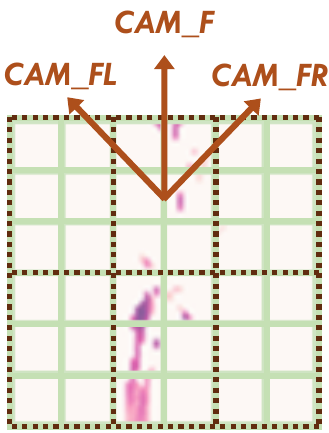} \hspace{-0.17cm}}
	\subfloat[Polar A \hspace{-0.15cm}]{
		\label{window(c)}
		\includegraphics[width=0.205\linewidth]{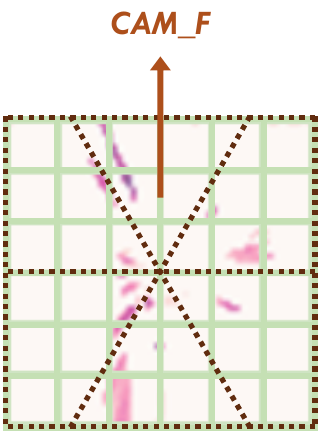} \hspace{-0.17cm}}
		\subfloat[Polar B \hspace{1cm}]{
		
		\label{window(d)}
		\includegraphics[width=0.286\linewidth]{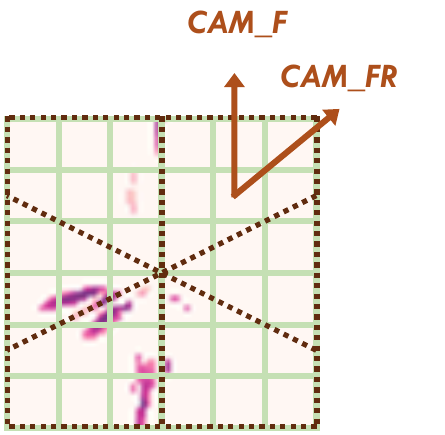} \hspace{-0.17cm}}
	\caption{\textbf{BEV Window Division.} The four window divisions are accomplished on BEV feature. Instead of listing each view group $G$ in detail, only the calculation of one window is given in the figure. Among them, (a) and (b) are rectangular divisions, and each window is interacted with three image views. (c) and (d) treats the BEV plane as in the polar coordinate system and divides it, whose $G$ includes one or two image views, respectively. For some methods, the number of grids in different windows is not equal, so we perform the zero padding additionally for batch operation.}
	\label{windows}
\end{figure}

To perform cross-attention in VA, BEV needs to be divided into multiple windows, and various partitioning methods could be considered. For choosing the optimal partitioning method as possible to balance the performance and computational cost, we have fully experimented with 4 different strategies in Fig. \ref{windows}. Before comparing the results experimentally, we first conduct a theoretical analysis of these strategies, calculating the computational reduction they bring compared to global attention:

\begin{equation}
\label{eq:7}
\begin{aligned}
    C_g&=O(6H_sW_sH_bW_b C)&=O(24576H_sW_s C)\\
    C_a&=O(4 \cdot 3H_sW_s \cdot \frac{H_b}{2} \cdot \frac{W_b}{2} \cdot C)&=O(12288H_sW_sC)\notag
\end{aligned}
\end{equation}    
\begin{equation}
\label{eq:8}
\begin{aligned}    
    C_b&=O(6 \cdot 3H_sW_s \cdot \frac{H_b}{2} \cdot \lceil \frac{W_b}{3} \rceil \cdot C)&=O(12672H_sW_sC)\\
    C_c&\approx O(6 \cdot H_sW_s \cdot \lceil S \rceil \cdot C)&=O(4374H_sW_sC)\\
    C_d&\approx O(6 \cdot 2H_sW_s \cdot \lceil S \rceil \cdot C)&=O(8748H_sW_sC),\notag
\end{aligned}
\end{equation}

\begin{table*}[htb]
    \caption{\textbf{Different attention types.} Global means vanilla multi-head attention. At an input resolution of $1600 \times 900$, only the experiments with global attention or (a) are performed, which is the chosen method according to the smaller resolution. \ding{55} means that the cost of global attention at a large resolution is too high and exceeds the memory limit.
}
	\centering
	\setlength{\tabcolsep}{1.3mm}{
		\linespread{1.1}\selectfont
		\begin{tabular}{c|ccccc|ccc}  \toprule
			Size      &Global &Rec 2 $\times$ 2 &Rec 2 $\times$ 3 &Polar A &Polar B &NDS$\uparrow$  &FLOPs$\downarrow$ &Memory$\downarrow$\\ 
			\hline
			\multirow{5}{*}{1600 $\times$ 640}
			&\checkmark & & & &   &0.434 &69.2G &19.0G\\
			& &\checkmark & & &   &\textbf{0.444} &27.4G &16.8G\\
			& & &\checkmark & &   &0.432 &25.5G &17.0G\\
			& & & &\checkmark &   &0.431 &\textbf{14.9G} &\textbf{15.8G}\\
			& & & & &\checkmark   &0.439 &20.6G &16.4G\\
			\hline            
		    \multirow{2}{*}{1600 $\times$ 900}  
		    &\checkmark & & & &   &\ding{55} &87.7G &\ding{55}\\
			& &\checkmark & & &   &\textbf{0.455} &\textbf{35.8G} &\textbf{23.0G}\\
			\hline
	\end{tabular}}
    \label{windowtb}
\end{table*}

where $H_b=W_b=64$ and $S=\frac{1}{2}\cdot\frac{H_b}{2}\left[W_b - \frac{H_b}{2\sqrt{3}}\right]$, the area of the largest divided part. It can be seen that (c) and (d) could contribute lower computational cost among all the methods, but in fact, they are more complicated and depend on the prior of the view, different from the rectangular windows. (a) and (b) have similar effects and reduce the computation by nearly half compared to the vanilla multi-head attention. 

The experimental results are given in Fig. \ref{windowtb}, which contains the comparison of two input image sizes under different attention types in terms of detection scores, computation, memory. For $1600 \times 640$ input, improvements of memory and FLOPs are basically consistent with the theoretical analysis, where (a) is slightly worse than (b) in Flops because it consider the calculation of the whole transformer. Although (c) performs the best on computational cost reduction, the NDS score is lower than other methods ($-1.3\%$ vs (a)) due to its insufficient represtation learning, only using one view. In contrast, (a) and (d) both achieve a good balance, whose NDS is better than global attention by about $1.0\%$ and $0.5\%$, along with sufficient cost reduction. Especially, all of our attention variants could achieve the comparable or better performance than origin attention in detection, demonstrating that VA effectively discards the redundant feature, promotes interactions within the effective region, and even improves the ability to learn BEV representation.

Focusing on detection performance, we finally choose the partitioning method of (a) and conduct the experiment with larger resolution $1600 \times 900$, which shows that the computational cost optimization is sufficient compared to global attention. 

In summary, VA, as our proposed novel attention variant, improves the original global attention cost and also replaces the unstable geometry-guidance attention. It is robust to camera parameters variation, ensuring sufficient receptive field and realizing the stable learning of potential view transformation.

\subsubsection{Noisy Extrinsics Analysis}
\label{Noisy Extrinsics Analysis}

\begin{figure}[htb]
	\centering  
	\subfloat{
		\includegraphics[width=0.9\linewidth ]{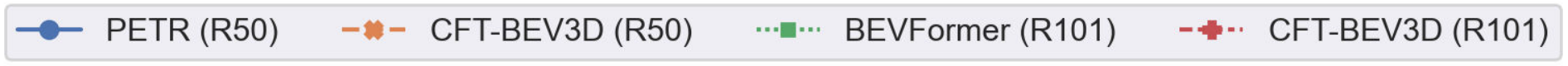}}\\
	\subfloat{
		\includegraphics[width=0.7\linewidth]{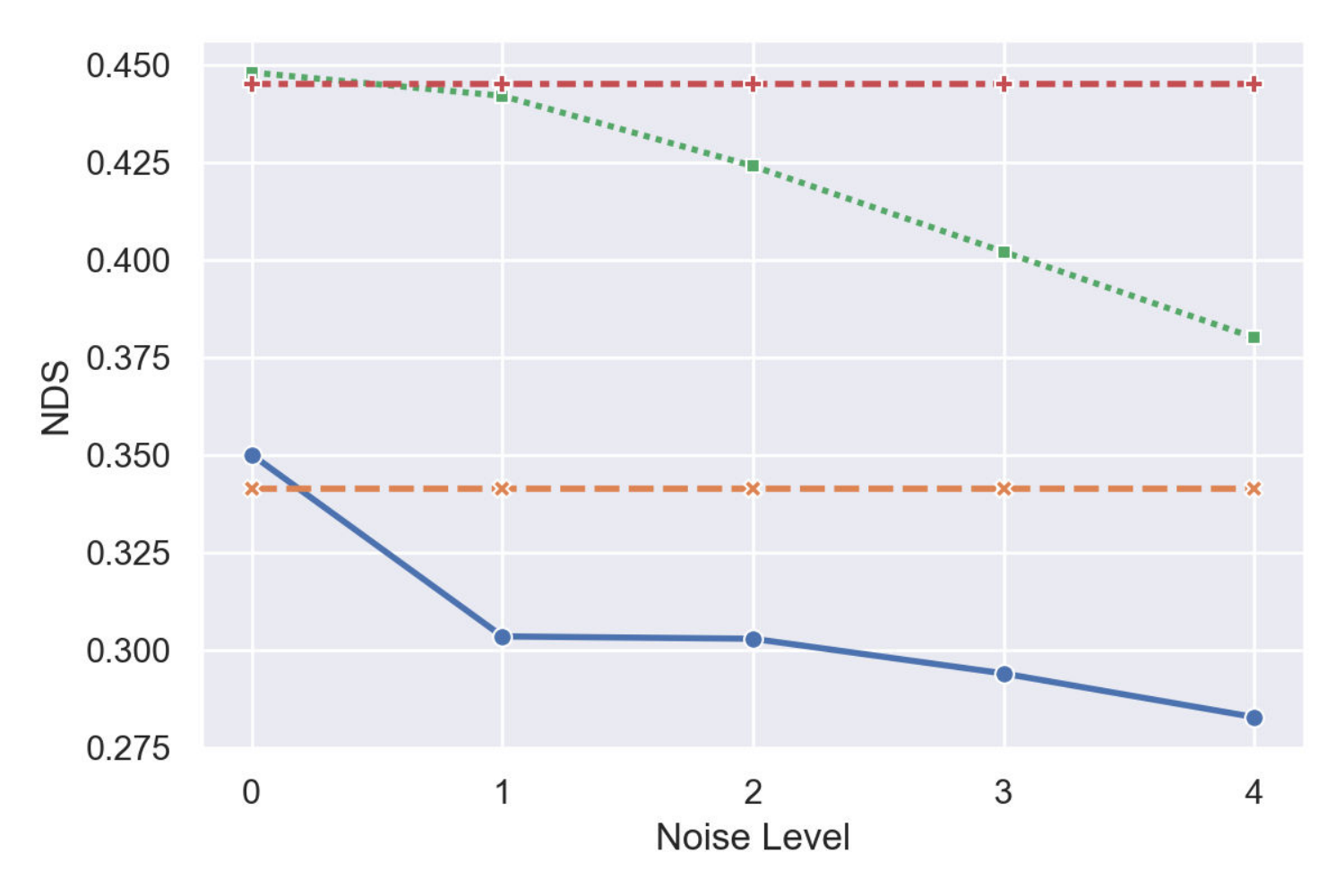}}
	\caption{\textbf{The performance of different models under noisy extrinsics.} The implementation of the noises is the same as \cite{Li2022BEVFormerLB}. The PETR (R50) is evaluated on the official pre-trained model, and our CFT-BEV3D (R50) is trained with the aligned common configuration.}
\label{noise}
\end{figure}

To analyze the impact of camera parameters on BEV representation learning, we adopt the extrinsics noises in \cite{Li2022BEVFormerLB}. It is applied on PETR, BEVFormer and our CFT-BEV3D to evaluate the trained model as shown in the results in Fig. \ref{noise}. Under the accurate extrinsics, the NDS score of CFT-BEV3D is slightly lower compared with other models of the same configuration. However, it is naturally undisturbed by the noise level variation with stable results. Even with a small perturbation of the parameters, BEVformer drops by $0.6\% - 6.8\%$ NDS and PETR drops by $4.7\% - 7.6\%$, but our method is not vulnerable to these perturbations and shows an evident improvement. The above degradation issues and results caused by the unstable camera parameters illustrate the importance of calibration free model, and confirms that competitive results can be obtained without relying on parameters guidance, whose stable performance is also more contributing to further expansion and application.

\subsection{Visualization}

\begin{figure}[htb]
	\centering  
	\subfloat[]{
	    \label{attention(a)}
		\includegraphics[width=0.2808\linewidth]{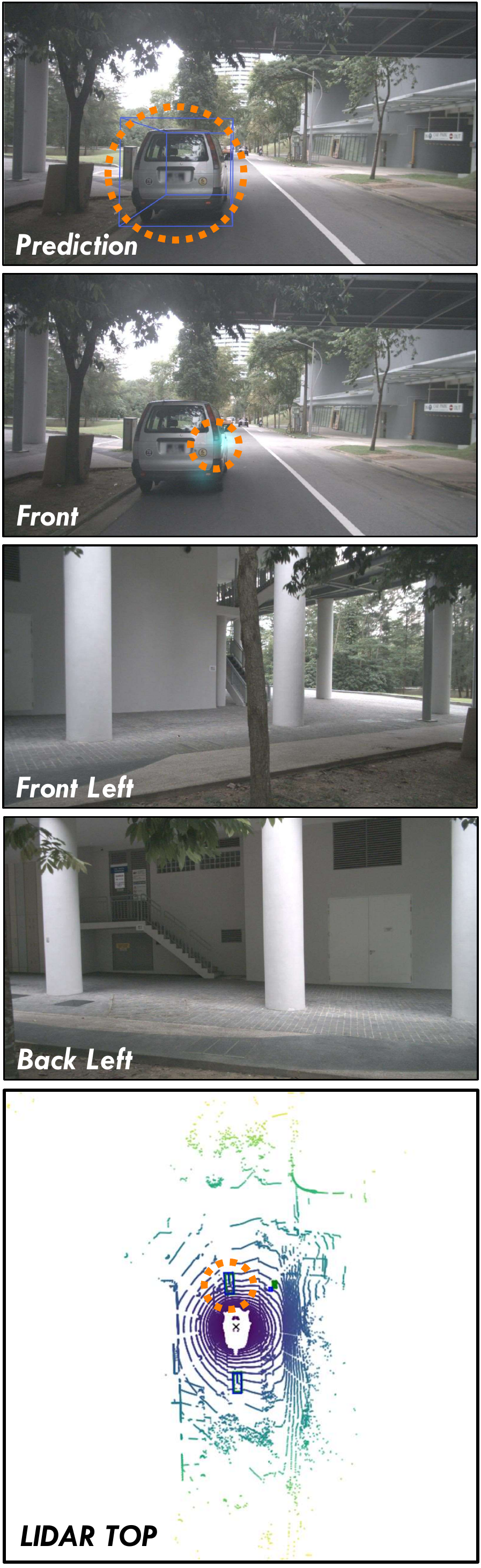}}
	\subfloat[]{
	    \label{attention(b)}
		\includegraphics[width=0.2807\linewidth]{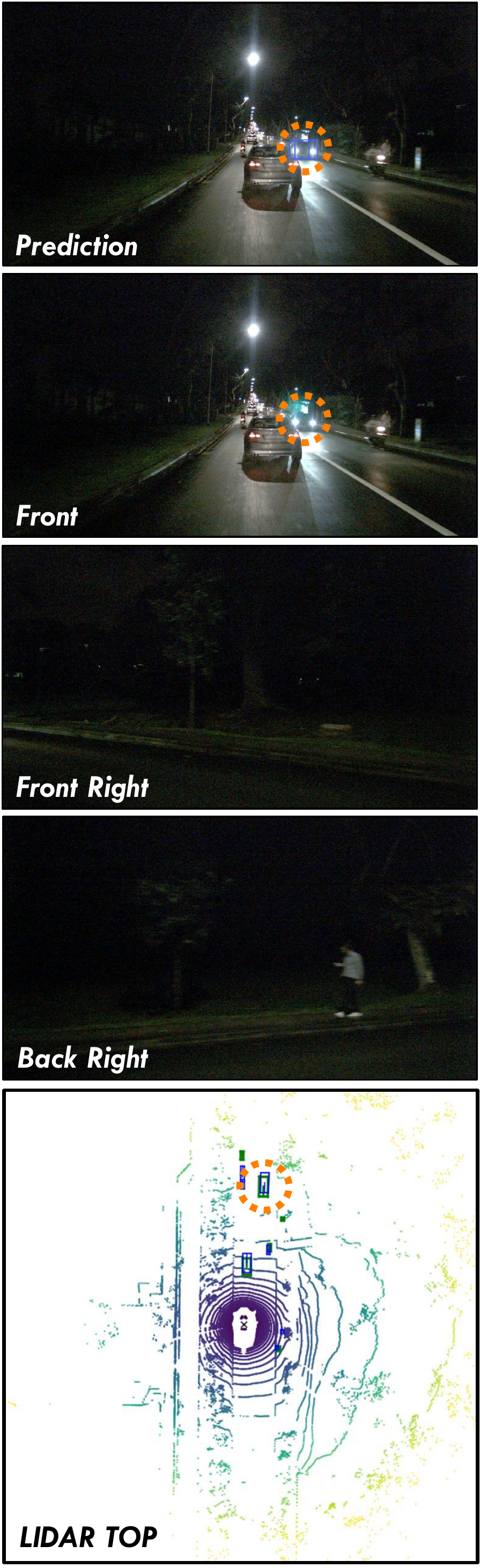}}
	\subfloat[]{
	    \label{attention(c)}
		\includegraphics[width=0.2818\linewidth]{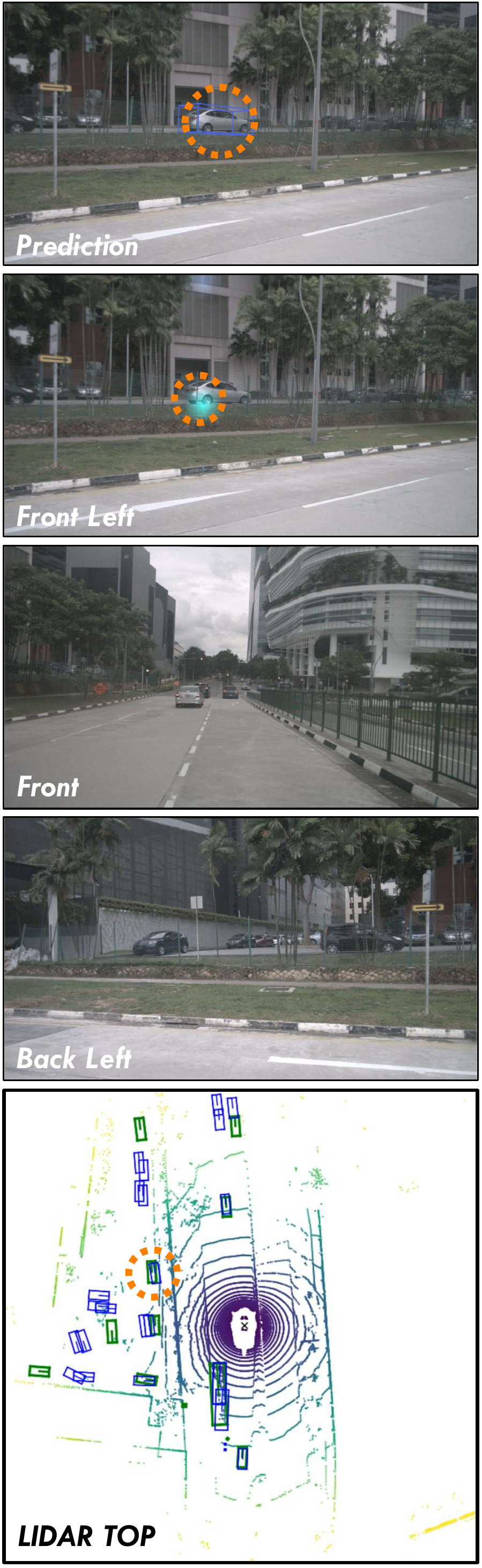}}
	\caption{\textbf{The attention weight maps of three views in the view group.} Each column shows a sample, in which the first row shows the object for visualization and its detection results. Three objects are located in the FL, FR, and FL window of the BEV, respectively.}
	\label{figure_attn}
\end{figure}

For the qualitative results, we show the visualization of detection results in Sec. \ref{Visualizations} and attention in Fig. \ref{figure_attn}. Specifically, we select the center points of objects with high confidence in the BEV plane, and visualize the bounding boxes predicted based on them. Then for a specific box, as the sample circled on the $LIDAR\_TOP$, we can visual the attention map calculated on each view in the corresponding view group. Three examples in Fig. \ref{figure_attn} show the results with different view groups. To further clarify the implicit enhancement of 3D information by PA, we choose the enhanced implicit design for the embedding. In fact, our model exhibits high responses to object regions, and are generally unresponsive in views without objects as shown in our samples. Also, VA effectively preserves the information, accelerates the convergence, and learns the transformation between image views and BEV without any geometry guidance.

\section{Conclusions}

In this paper, we present an approach for learning BEV representation from image views via a completely camera calibration free view transformation. Our proposed CFT is the first parameters-free work, comparable to the state-of-the-art methods in 3D object detection, thanks to learning mapping implicitly with enhanced 3D information. Moreover, we propose an attention variant, which enables the model to reduce memory, computation cost and converge better. Unlike existing  CFT is naturally robust to camera parameters noise. In the near future, we will explore combining CFT with temporal or additional modalities.


\appendix

\section{Appendix}

\subsection{Visualizations}
\label{Visualizations}
\begin{figure*}[htb]
\centering
\includegraphics[width=\textwidth]{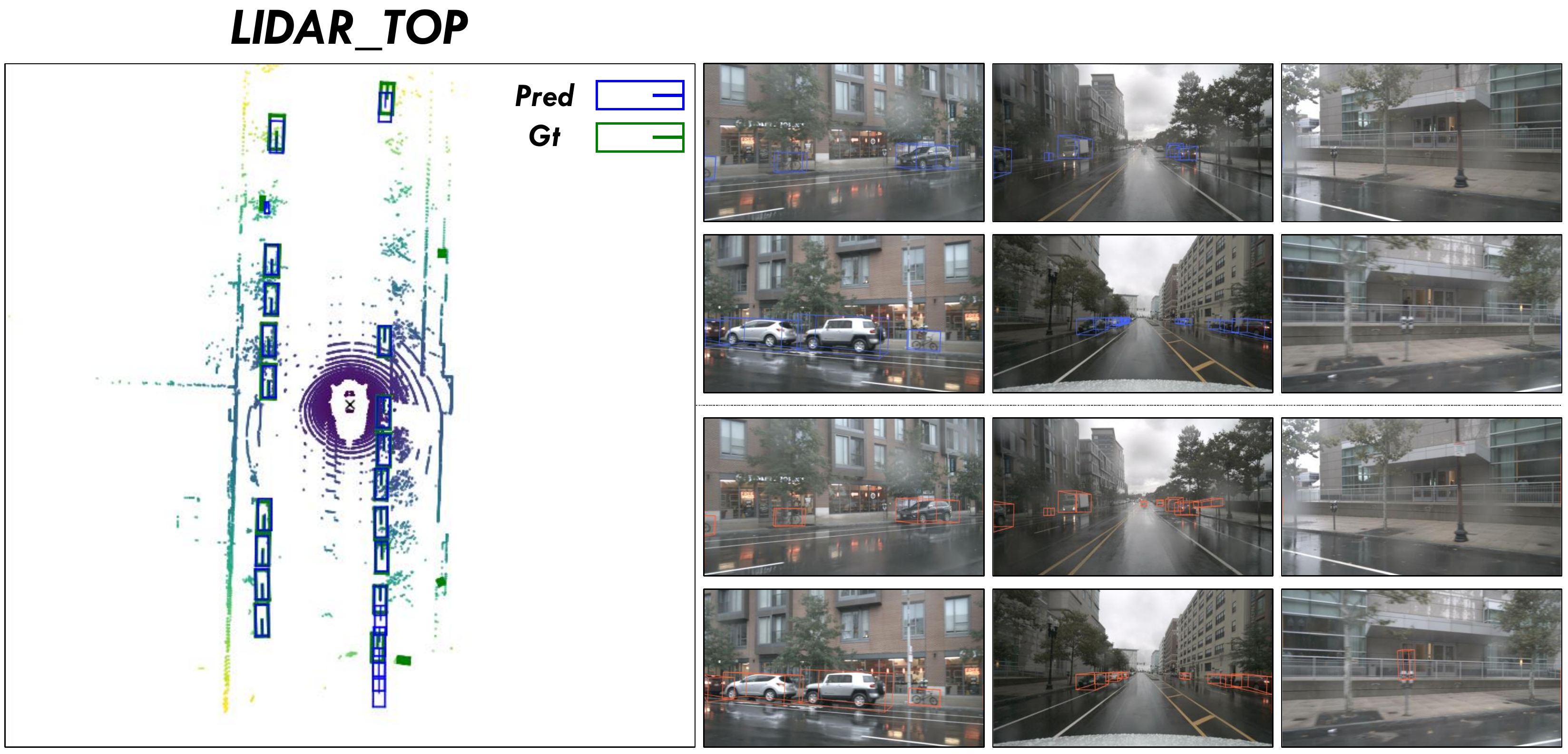}
\caption{\textbf{3D object detection results in BEV and image views.} The figure includes the visualization of 3D bounding boxes in different views. The blue boxes in the image views are the prediction results, and the orange boxes are the ground truth.}
\label{results}
\end{figure*}

We visualize our 3D object detection results in Fig. \ref{results}. The prediction results in BEV are close to the ground truth, indicating that our method performs well within the effective detection range. Since the BEV size we directly utilize for cross-attention is only $64$, which may be unfavorable for detection of some objects with small 2D size such as pedestrians.

\end{document}